\documentclass[letterpaper, 10 pt, conference]{ieeeconf}
\usepackage{amsmath,amsfonts}
\usepackage{algorithmic}
\usepackage{algorithm}
\usepackage{array}
\usepackage[caption=false,font=normalsize,labelfont=sf,textfont=sf]{subfig}
\usepackage{textcomp}
\usepackage{stfloats}
\usepackage{url}
\usepackage{verbatim}
\usepackage{graphicx}
\usepackage{cite}
\IEEEoverridecommandlockouts                              
\title{\LARGE \bf
T-ConSG: A LiDAR-Based Real-Time Temporal Construction Scene Graph for Dynamic and Evolving Environments
}
\author{Yilong Chen$^{1}$ and Yong K. Cho$^{1}$
\thanks{The authors are with the Department of Civil and Environmental Engineering, Georgia Institute of Technology, Atlanta, GA 30332, USA {(\tt\small e-mail: ychen3339@gatech.edu, yong.cho@ce.gatech.edu)}}%
}
\begin{document}
\maketitle
\thispagestyle{empty}
\pagestyle{empty}
\begin{abstract}
Dynamic and continuously evolving environments present significant challenges for robotic mapping and scene understanding, particularly on construction sites where moving agents, temporary obstacles, and persistent structural changes coexist. While dynamic and long-term SLAM methods can detect motion and environmental changes, their outputs are predominantly metric representations that provide limited structure for higher-level spatial and temporal reasoning. This paper presents a LiDAR-based real-time temporal construction scene graph (T-ConSG) that jointly represents free-space regions, static objects, dynamic objects, and environmental events with spatial-relational and event-participation edges. The proposed framework incrementally constructs the graph from LiDAR-derived geometric evidence, maintaining persistent free-space and static-object information while tracking dynamic objects. Changes in accumulated spatial evidence are further represented as explicit installation and removal events, allowing the graph to capture both transient dynamics and persistent environmental evolution. We demonstrate the proposed representation through three complementary application studies: dynamics-aware navigation in a bricklaying testbed, construction progress monitoring on the ConSLAM dataset, and online change detection in a simulated construction environment. The results demonstrate that the proposed T-ConSG provides a unified representation for organizing spatial and temporal information and supports robotic reasoning across environments containing both short-term dynamics and persistent changes.
\end{abstract}

\section{Introduction}
Construction environments present a particularly demanding setting for autonomous robots because their spatial configuration evolves continuously throughout project execution. Unlike structured industrial workplaces, construction sites contain incomplete structures, temporary facilities, relocated materials, moving workers and equipment, and work areas whose accessibility changes as construction progresses \cite{ parascho2023construction, chen2026localization}. These characteristics create substantial challenges for perception, navigation, and autonomous decision-making, while simultaneously motivating robotic systems that can improve productivity and reduce human exposure to hazardous tasks \cite{parascho2023construction, you2025construction}. Effective robotic operation therefore requires more than estimating the current geometry of the surroundings; robots must maintain spatial knowledge that distinguishes transient motion from persistent environmental changes and preserves how the workspace evolves over time.

Most SLAM systems were originally formulated around a predominantly static world, an assumption that becomes restrictive when robots operate amid moving agents and persistent environmental changes. Dynamic SLAM has progressively relaxed this assumption by rejecting dynamic measurements, tracking independently moving objects, or reconstructing dynamic scenes, while long-term and change-aware approaches further address changes that become apparent when previously mapped areas are revisited \cite{Schmid_2023,schmid2024khronos,sh-ch15-dyndef}. Nevertheless, these methods are primarily concerned with maintaining localization, metric reconstruction, object states, or geometric change estimates. For autonomous operation in evolving construction environments, such metric evidence would be more useful if it could also be organized into persistent entities and explicit events that support higher-level spatial and temporal reasoning.

Scene graphs provide a complementary abstraction by organizing environments as entities and relationships rather than only metric geometry. Recent 3D scene graphs have demonstrated the value of hierarchical representations for robotic mapping, planning, and spatial reasoning, while emerging 4D formulations extend this concept to entities and relationships that evolve over time \cite{armeni20193d, rosinol20203d, hughes2022hydra, schmid2024khronos, rotondi20263d}. However, temporal scene-graph research remains comparatively less mature, particularly in representing how persistent environmental changes modify both spatial entities and their relationships. Moreover, free space is often treated primarily as geometric support for navigation rather than as persistent spatial evidence coupled with scene changes. This motivates a 4D representation that jointly structures free space, static objects, dynamic objects, and explicit environmental events.

We propose T-ConSG, a LiDAR-based, real-time temporal construction scene graph that provides a structured representation of both the spatial state and temporal evolution of an evolving construction environment. To the best of our knowledge, T-ConSG is the first LiDAR-based scene-graph framework to jointly represent free-space regions, static structures, tracked dynamic objects, and explicit temporal events for construction-oriented robotic applications. The graph connects environmental states and detected changes through two complementary edge families: spatial-relational edges describe the current spatial configuration and accessibility of graph entities, whereas event-participation edges associate temporal changes with the entities they affect. This representation distinguishes transient object motion from persistent structural changes while incrementally maintaining the current spatial state from LiDAR-derived geometric evidence, without requiring camera-based semantic perception. Accordingly, T-ConSG provides a compact, task-oriented relational abstraction for downstream reasoning. The main contributions of this work are threefold: (1) a temporal scene-graph representation that jointly models spatial states and environmental events in evolving environments; (2) a real-time, incremental LiDAR mapping framework that integrates dynamic-object tracking with the detection and graph-based representation of persistent temporal events; (3) the versatility of T-ConSG through three construction-oriented applications: dynamics-aware navigation, construction progress monitoring, and online change detection.
\section{Related Work}
\subsection{3D and 4D Scene Graphs}
3D scene graphs (3DSGs) have emerged as structured world representations that combine metric grounding with semantic entities and explicit relationships, providing a compact interface between perception, reasoning, and action \cite{armeni20193d,rosinol20203d,hughes2022hydra}. Recent formulations have extended early object- and room-centric graphs toward hierarchical online representations that incorporate places, agents, and metric geometry \cite{chang2023hydra, greve2024collaborative,schmid2024khronos}. However, extending these representations to continuously evolving environments remains an open problem. Rotondi et al. distinguish 4D scene graphs (4DSGs) by allowing graph components to vary over time and emphasize that effective representations must capture both short-term dynamics and long-term environmental changes \cite{rotondi20263d}. Existing approaches generally represent dynamics through graph snapshots\cite{wang2023trajectory}, observation-specific nodes \cite{schmid2024khronos}, or time-varying node and edge attributes \cite{greve2024collaborative,rosinol2021kimera}. While the latter is compact and well suited to tracking moving agents, it relies heavily on reliable temporal data association. Long-term approaches have begun maintaining changing environments and reconstructing historical scene states \cite{gorlo2026describe, nguyen2025react}, yet a general 4DSG should represent not only entity motion but also spatiotemporal relationships and higher-level temporal abstractions, which remain comparatively underexplored. This limitation is particularly important in unstructured environments such as construction sites, where moving workers and equipment coexist with persistent structural changes. Moreover, current research predominantly tracks entities over time rather than reasoning about the structural and relational changes induced by their evolution, while reliable change detection itself remains challenging. Although free or traversable space has previously been incorporated as graph nodes for planning \cite{rosinol20203d, hughes2022hydra, rosinol2021kimera}, its temporal interaction with dynamic and persistent objects has received considerably less attention. Motivated by these limitations, we develop a temporal scene graph that jointly represents accumulated free space, static structures, tracked dynamic objects, and explicit environmental events, thereby connecting persistent LiDAR mapping with higher-level temporal reasoning in evolving construction environments.
\subsection{Dynamic and Long-Term SLAM}
Classical SLAM commonly relies on scene rigidity, whereas real environments contain both continuously moving entities and persistent structural changes \cite{sh-ch15-dyndef}. Looper et al. distinguish short-term and long-term dynamics according to the relationship between scene change and observation rate, rather than absolute duration: motion observed continuously appears transient, while changes occurring outside the sensor's view appear abrupt when the region is revisited \cite{looper20233d}. Early dynamic SLAM approaches primarily protected state estimation by rejecting independently moving measurements, while subsequent methods explicitly track moving entities, including MVO \cite{judd2018multimotion}, VDO-SLAM \cite{zhang2020vdo}, DynaSLAM II \cite{bescos2021dynaslam}, and dense reconstruction approaches such as Co-Fusion \cite{runz2017co}. Dense LiDAR methods such as Dynablox \cite{Schmid_2023} instead exploit volumetric consistency to identify dynamic measurements without requiring predefined object classes. However, short-term tracking alone does not address environments whose geometry changes while unobserved. Long-term changes introduce substantially harder association problems and encompass object addition, relocation, and removal as well as structural modifications. Construction sites are explicitly identified as an example where successive observations may differ fundamentally \cite{sun2025nothing}. Change-detection methods therefore compare observations or maps across time, with volumetric occupancy or TSDF representations being particularly valuable because they distinguish unobserved space from observed free space, allowing newly occupied and newly cleared regions to be identified \cite{fehr2017tsdf, rowell2024lista}. Recent change-aware systems, including Panoptic Multi-TSDFs \cite{schmid2022panoptic}, extend this reasoning toward online operation, but continuously separating genuine changes from localization and measurement errors remains challenging. Unified approaches such as Khronos \cite{schmid2024khronos} further combine short-term motion with long-term change estimation, which targets dense reconstruction of the state of a dynamic scene over time. Unlike Khronos, T-ConSG converts construction-relevant state transitions into explicit event nodes and updates the associated spatial and accessibility relations, supporting dynamics-aware navigation, progress monitoring, and structural-change detection.
\section{Methodology}
\begin{figure}[!t]
\vspace*{4pt}
\centering
\includegraphics[width=\linewidth]{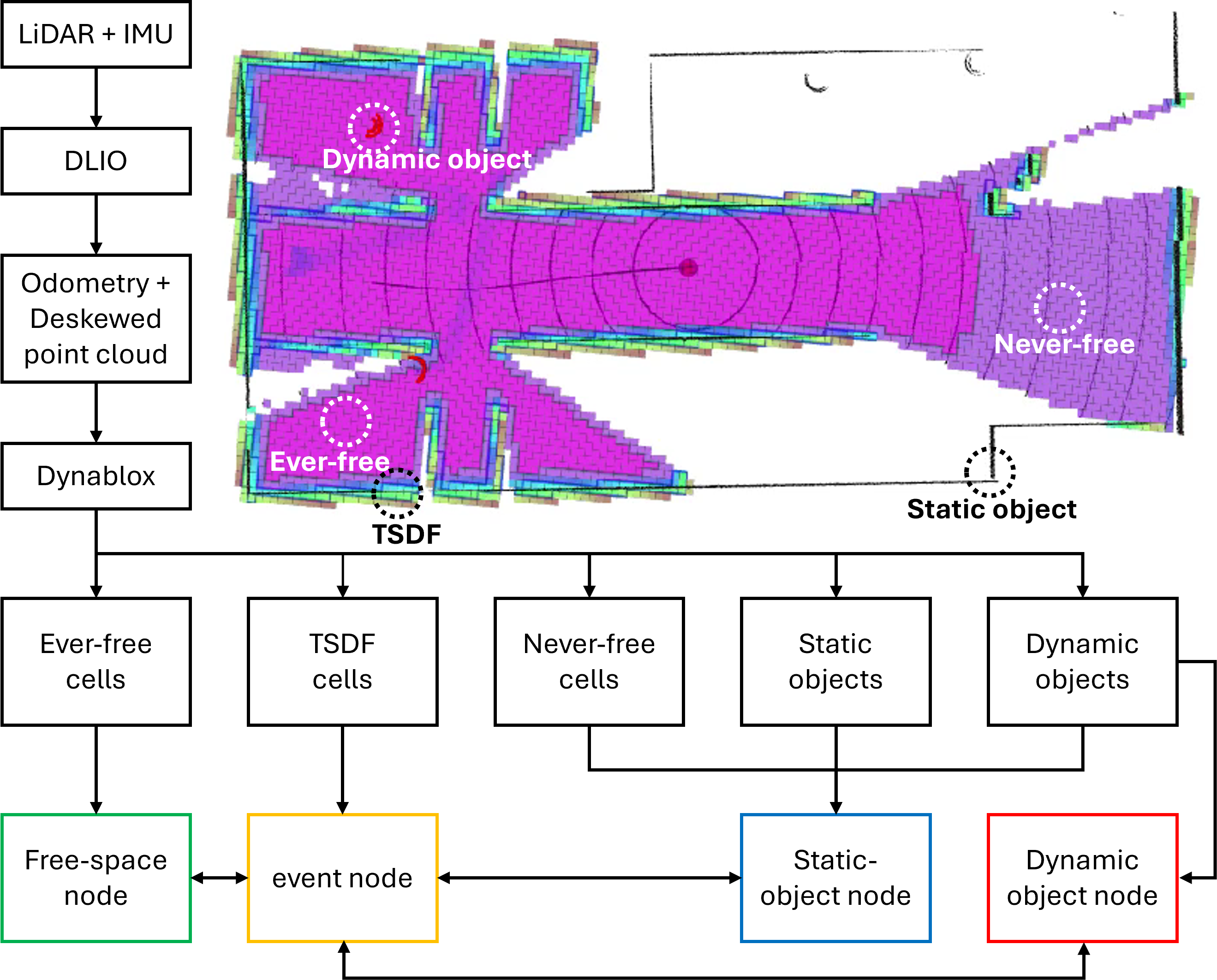}
\caption{Overall workflow.}
\label{fig_workflow}
\end{figure}

This section describes how the proposed scene graph is constructed (Fig.~\ref{fig_workflow}). We employ DLIO \cite{chen2022direct} as the default odometry backend because it provided more stable pose estimates and greater robustness to the rapid, irregular motion of the quadruped platform in our experiments. The outputs of Dynablox, including the ever-free map, TSDF map, never-free map, static objects, and dynamic objects, are jointly used for constructing the T-ConSG. 

As illustrated in Fig.~\ref{fig_4dsg}, the proposed T-ConSG \(G_t\) at time \(t\), defined in \eqref{eqn_graph}, integrates persistent vertices \(V_t\) \eqref{eqn_vertex} including free-space \(V^{F}_t\) and static-object \(V^{S}_t\) representations, tracked dynamic objects \(V^{D}_t\), and detected environmental events \(V^{E}_t\) within a unified spatiotemporal representation. Its edge set \(E_t\) \eqref{eqn_edge} encodes both spatial relationships \(E_t^{R}\)  among world-state nodes and temporal participation relationships \(E_t^{T}\) that associate installation, removal, appearance, disappearance, and observation events with the affected graph entities. Although the input consists of three-dimensional LiDAR measurements and volumetric mapping outputs, the present implementation projects the relevant observations onto a horizontal plane to support ground-robot navigation and structural-change detection with reduced computational complexity. The proposed node, relation, and event definitions are not restricted to this planar representation and can be generalized to three-dimensional voxel components, although efficient volumetric connectivity and boundary extraction remain subjects for future implementation.
\begin{align}
    G_t &= (V_t,E_t) \label{eqn_graph}\\
    V_t &= V^{F}_t\cup V^{D}_t\cup V^{S}_t\cup V^{E}_t \label{eqn_vertex} \\
    E_t & =E^{R}_t\cup E^{T}_t \label{eqn_edge}
\end{align}
\subsection{Free-Space Node}
The free-space node \(V^{F}_t\) maintains a persistent two-dimensional representation of historically observed free space \(F_t\) from the horizontal ever-free voxel slice generated by the LiDAR mapping system. Each observed voxel center is quantized into an integer grid cell according to the TSDF voxel resolution, and the persistent free-space set \(F_t\) is updated through addition-only fusion of \(F_{t-1}\) with \(Er_t\), where \(Er_t\) denotes the ever-free cells observed at time \(t\) as defined in \eqref{eqn_free}. Newly inserted cells are associated through four-neighbor connectivity, while a union-find structure incrementally creates and merges connected free-space components without repeatedly processing unchanged regions. For each affected component, its polygonal boundary is recovered by constructing the four directed edges of every constituent cell and cancelling shared edges between adjacent cells. The remaining edges are traced into closed loops and classified according to their orientation, producing an outer boundary together with zero or more inner boundaries. These inner boundaries denote areas excluded from the confirmed historical free space and may correspond to occupied or insufficiently observed regions; therefore, they are not interpreted as obstacles without additional evidence. The resulting cell indices and polygonal boundaries constitute the geometry of the free-space nodes and support subsequent spatial-overlap and temporal-change reasoning. Although the representation is accumulated over time, cells associated with a confirmed installation event  \(\mathcal{I}_t\) from event node \(V^{E}_t\) are explicitly removed, after which the affected connectivity and polygon geometry are reconstructed to reflect the newly occupied region.
\begin{align}
    F_t &= (F_{t-1}\cup Er_t \label{eqn_free}) \setminus \mathcal{I}_t
\end{align}
\subsection{Dynamic-Object Node}
\begin{figure}[!t]
\vspace*{4pt}
\centering
\includegraphics[width=\linewidth]{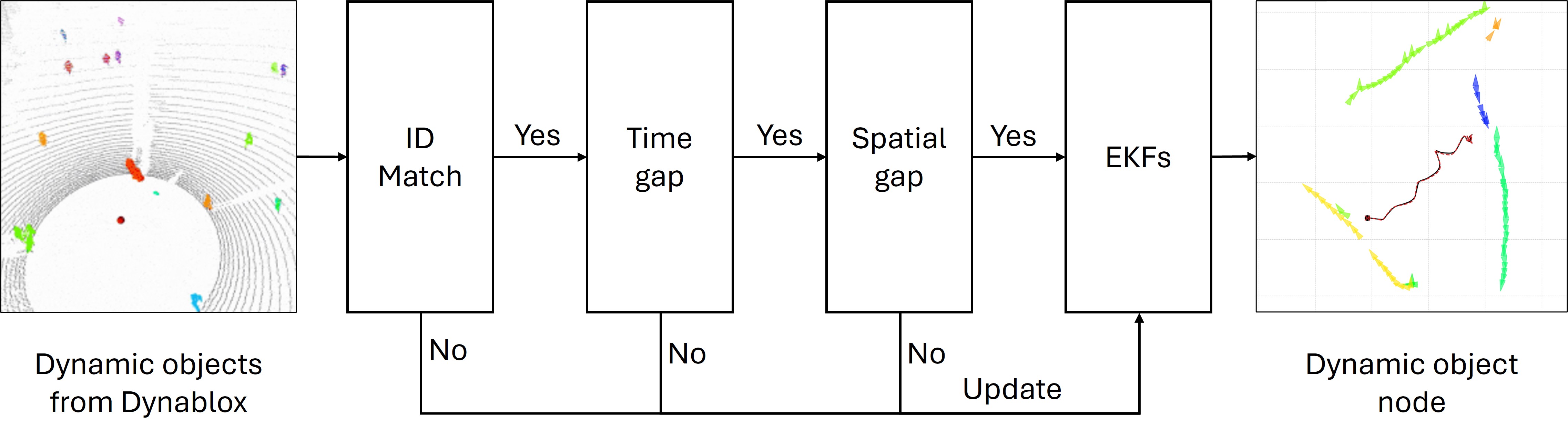}
\caption{Dynamic-object tracking and association.}
\label{fig_dynamic_object_tracking}
\end{figure}

Dynablox \cite{Schmid_2023} provides frame-level dynamic-object detections from LiDAR observations. However, additional temporal processing is required before these detections can be incorporated into the scene graph. In particular, conservative free-space inference and sensor noise may introduce discontinuities in the detected segments. To obtain temporally consistent object states, we augment Dynablox with an extended Kalman filter (EKF) tracking module, as shown in Fig.~\ref{fig_dynamic_object_tracking}. Each incoming detection is compared with the active tracks stored in a set of active EKF tracks using its object identifier, spatial displacement, and observation interval. The detection is assigned to an existing track only if the identifier is consistent and the spatial and temporal differences remain within their respective thresholds. Following a successful association, the EKF propagates the tracked state and corrects it using the current observation. If these association criteria are not satisfied, the previous track is terminated and a new EKF instance is initialized for the detection. Through this process, isolated frame-level detections are converted into continuous trajectories with estimated motion states, including object velocity. These tracked object states are maintained in the dynamic-object node \(V^{D}_t\). The spatial and temporal association thresholds are empirically set to fixed values.
\subsection{Static-Object Node}
Static object nodes \(V^{S}_t\) are incrementally constructed on a two-dimensional grid \(S_t\) by combining the never-free state of the volumetric map with object-level motion classification. At each update, the never-free voxel slice \(N_t\) and object-level static detections \(O^{\text{static}}_t\) are synchronized and projected onto a common horizontal grid. A cell is considered a static-object candidate when it is both never-free and contains at least one object-level static observation. To reduce contamination from moving objects and motion-classification uncertainty near object boundaries, cells associated with object-level dynamic detections \(O^{\text{dynamic}}_t\) are spatially dilated and removed from the candidate set. The resulting static cells are accumulated into a persistent map \(S_t\), such that repeated observations reinforce the spatial extent of static structures while previously detected cells \(S_{t-1}\) are retained \eqref{eqn_static_object}. Following the same connectivity and polygonization procedure as the free-space node, newly observed static cells are incrementally merged into connected components and represented as polygons, while their cell indices are retained for efficient map comparison and event detection. The static map is accumulated in an addition-only manner during normal observation, but cells can be removed through explicit removal events \(\mathcal{R}_t\) from event node \(V^{E}_t\), allowing persistent structural changes to be reflected in the scene graph.
\begin{align}
S_t &= (S_{t-1}\cup (N_t\cap O^{\text{static}}_t \setminus O^{\text{dynamic}}_t)) \setminus \mathcal{R}_t \label{eqn_static_object}
\end{align}
\subsection{Event Node}
The event nodes \(V^{E}_t\) represent long-term structural changes by identifying conflicts between the accumulated free-space \(F_t\) and static-object maps \(S_t\). Since both spatial maps retain previously observed cells, their intersection \(C_t\) represents locations that have been classified as both free and static over time and are therefore treated as potential change regions \eqref{eqn_change}. For each conflicting grid cell, a set of cells satisfying an occupancy criterion \(\mathrm{TSDF}_t\) is used to determine the direction of change: cells supported by TSDF measurements within a predefined range are considered installation candidates \(I_t^{\mathrm{cand}}\) \eqref{eqn_installation_candidate}, whereas conflicting cells without such support are considered removal candidates \(R_t^{\mathrm{cand}}\) \eqref{eqn_removal_candidate}. To suppress transient inconsistencies and sensor noise, a signed temporal counter \(q_t(c)\) is maintained independently for each cell \(c\), where consecutive installation observations increase the counter and consecutive removal observations decrease it; a change in candidate type resets the counter direction. A structural change is accepted only after the counter  exceeds a predefined persistence threshold \(N_e\). In addition, the set of confirmed candidate cells is required to remain spatially unchanged for several consecutive updates before an installation event \(\mathcal{I}_t\) \eqref{eqn_installation_event} or removal event \(\mathcal{R}_t\) \eqref{eqn_removal_event} is published, providing a second level of temporal stabilization. The confirmed cells are subsequently grouped using four-neighbor connectivity, and the boundaries of each connected component are converted into event polygons. The resulting event node records either an installation or removal event together with its affected polygonal region and grid-cell indices, which can subsequently be used to update the corresponding free-space or static-object representation. In addition to the long-term change events, \(V^{E}_t\) could also record other essential events, such as the appearance and disappearance of dynamic objects, or the first observation of a static object.
\begin{align}
    C_t &= F_t\cap S_t \label{eqn_change} \\
    I_t^{\mathrm{cand}} &= C_t \cap \mathrm{TSDF}_t \label{eqn_installation_candidate}\\
    R_t^{\mathrm{cand}} &= C_t \setminus \mathrm{TSDF}_t \label{eqn_removal_candidate}\\
    \mathcal{I}_t &= \left\{ c \in C_t \mid q_t(c) \geq N_e \right\} \label{eqn_installation_event}\\
    \mathcal{R}_t &= \left\{ c \in C_t \mid q_t(c) \leq -N_e \right\} \label{eqn_removal_event}
\end{align}
\begin{figure}[!t]
\vspace*{4pt}
\centering
\includegraphics[width=\linewidth]{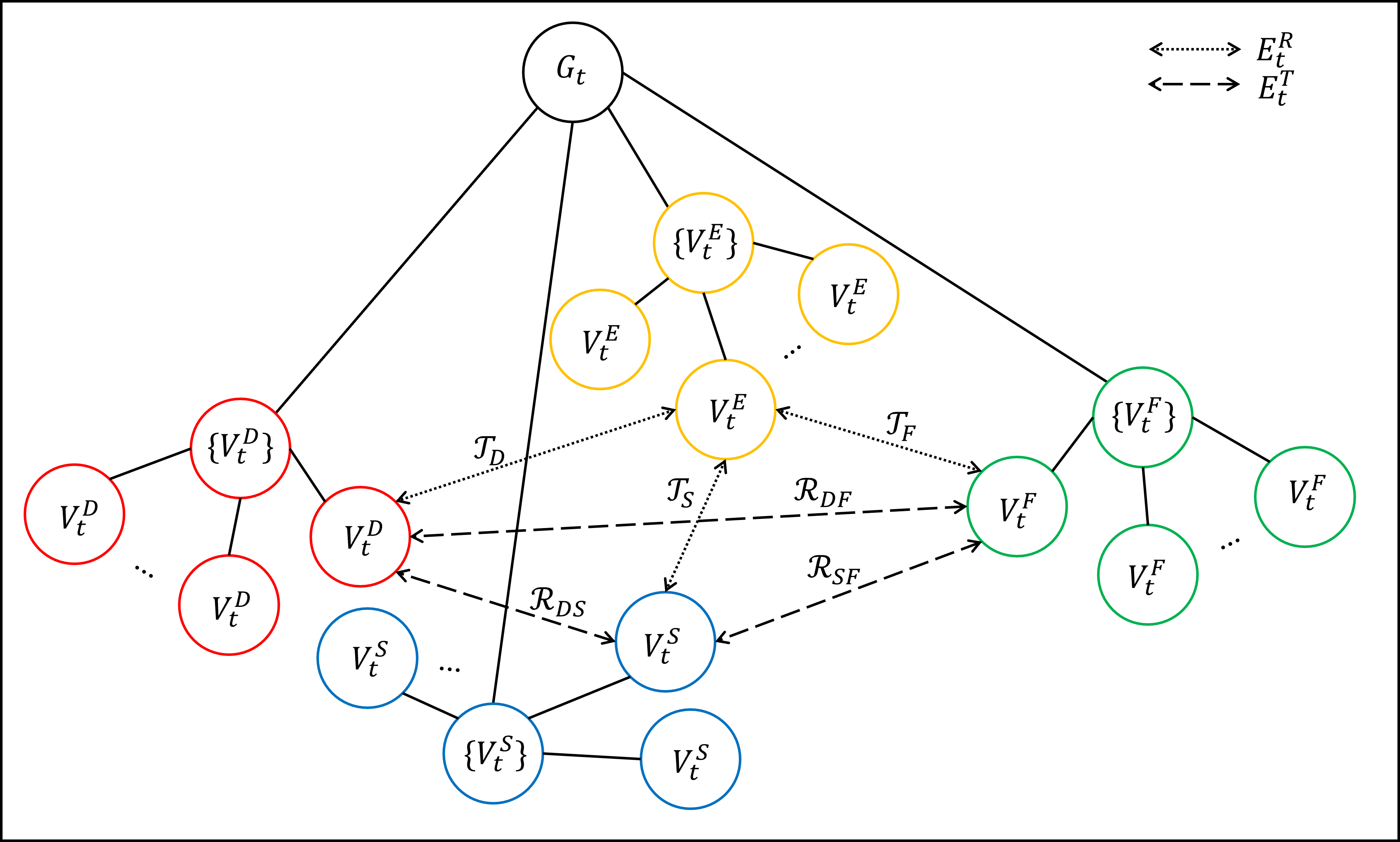}
\caption{Temporal construction scene graph.}
\label{fig_4dsg}
\end{figure}
\subsection{Graph Edge Construction}
The edges \(E_t\) of the proposed T-ConSG \(G_t\) are divided into two families: spatial-relational edges \(E^{R}_t\) \eqref{eqn_spatial_relational} and event-participation edges \(E^{T}_t\) \eqref{eqn_event_participation}, as shown in Fig.~\ref{fig_4dsg}. The nodes constructing these edges include the free-space nodes \(V^{F}_t\), dynamic-object nodes \(V^{D}_t\), static-object nodes \(V^{S}_t\), and event nodes \(V^{E}_t\).
\begin{align}
    E^{R}_t \subseteq {}&\left(V^{D}_t\times V^{F}_t\times\mathcal{R}_{DF}\right)\cup\left(V^{S}_t\times V^{F}_t\times\mathcal{R}_{SF}\right) \nonumber\\&\cup\left(V^{D}_t\times V^{S}_t\times\mathcal{R}_{DS}\right) \label{eqn_spatial_relational}\\
    E^{T}_t \subseteq {}&\left(V^{E}_t\times V^{D}_t\times\mathcal{T}_{D}\right)\cup\left(V^{E}_t\times V^{S}_t\times\mathcal{T}_{S}\right) \nonumber\\&\cup\left(V^{E}_t\times V^{F}_t\times\mathcal{T}_{F}\right) \label{eqn_event_participation}
\end{align}
{\textbf{Spatial-Relational Edges:}} The spatial-relational edge set \(E_t^{R}\) connects the world-state nodes \(V_t^{F}\cup V_t^{D}\cup V_t^{S}\) and describes their spatial or accessibility relationships. In particular, a dynamic-object node \(V_t^{D}\) and a free-space node \(V_t^{F}\) are connected by the directional relations \(\mathcal{R}_{DF} = \{\texttt{inside},\,\texttt{include}\}\), indicating that the dynamic object is located within the corresponding connected free-space region or, inversely, that the free-space region includes that object. A static-object node \(V_t^{S}\) is linked to a free-space node \(V_t^{F}\) through a connected relation \(\mathcal{R}_{SF} = \{\texttt{connected}\}\) when the object is spatially associated with and accessible from that free-space component. Based on these relations, a dynamic-object node \(V_t^{D}\) and a static-object node \(V_t^{S}\) can further be associated by the directional relations \(\mathcal{R}_{DS} = \{\texttt{can\_access},\,\texttt{can\_be\_accessed}\}\). Specifically, if a dynamic object is inside a free-space node and a static object is connected to the same free-space node, the dynamic object is considered capable of accessing that static object. The edge labels in Fig. ~\ref{fig_4dsg} therefore represent the two spatial directions of the same relation.

{\textbf{Event-Participation Edges:}} The event-participation edge set \(E_t^{T}\) connects event nodes \(V_t^{E}\) with the world-state nodes \(V_t^{F}\cup V_t^{D}\cup V_t^{S}\) and captures temporally meaningful changes observed in the environment. In contrast to the spatial-relational edges, the labels on an event edge in Fig. ~\ref{fig_4dsg} denote different event types, rather than opposite directions of one spatial relation. For dynamic-object nodes \(V_t^{D}\), the corresponding events \(\mathcal{T}_{D} = \{\texttt{appear},\,\texttt{disappear}\}\) describe whether an object appears or disappears from observation. For static-object nodes \(V_t^{S}\), the event relation records \(\mathcal{T}_{S} = \{\texttt{observed},\,\texttt{removed}\}\), indicating whether a static object is observed or removed, while for free-space nodes \(V_t^{F}\) it records \(\mathcal{T}_{F} = \{\texttt{observed},\,\texttt{installed}\}\), showing whether a region is observed as free or is affected by an installed object. In this way, \(E_t^{T}\) explicitly associates a detected temporal change with the graph entity affected by that change, while the event node itself preserves the event identity and timestamp. The edge set \(E_t^{R}\) describes the current spatial configuration of the environment, whereas \(E_t^{T}\) records how that configuration evolves over time, allowing the proposed graph to represent both the instantaneous scene state and its temporal development.
\section{Application and Discussion}
As no single available dataset covers all capabilities of T-ConSG, we use three complementary construction-related datasets to examine its construction and demonstrate its use in navigation, construction progress monitoring, online change detection. In all experiments, T-ConSG ran at a real-time processing rate on the robot's onboard CPU.
\subsection{Dynamics-Aware Navigation}
\begin{figure}[!t]
\centering
\subfloat[]{\includegraphics[width=0.41\linewidth]{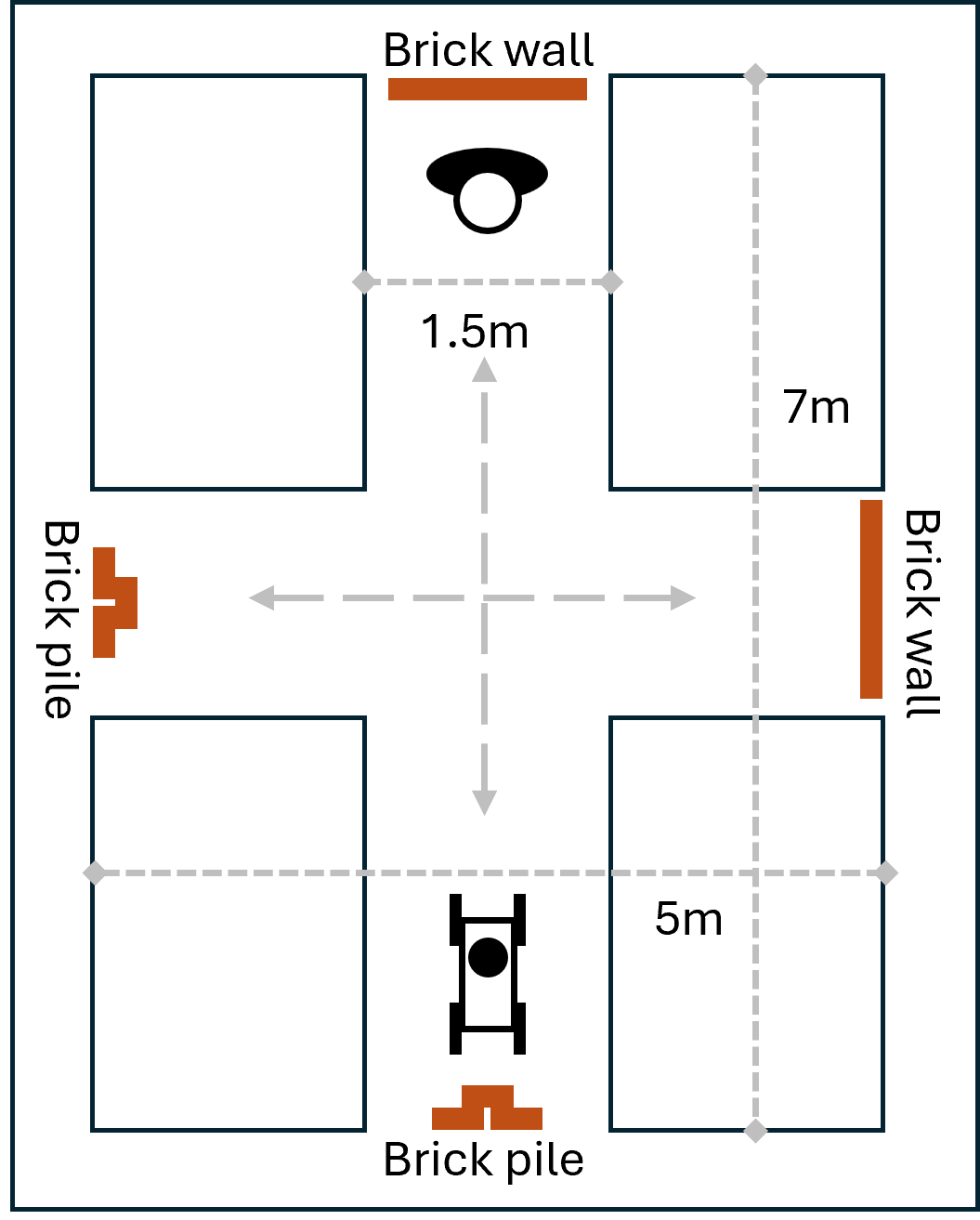}%
\label{fig_bricklaying_layout}}
\hfil
\subfloat[]{\includegraphics[width=0.56\linewidth]{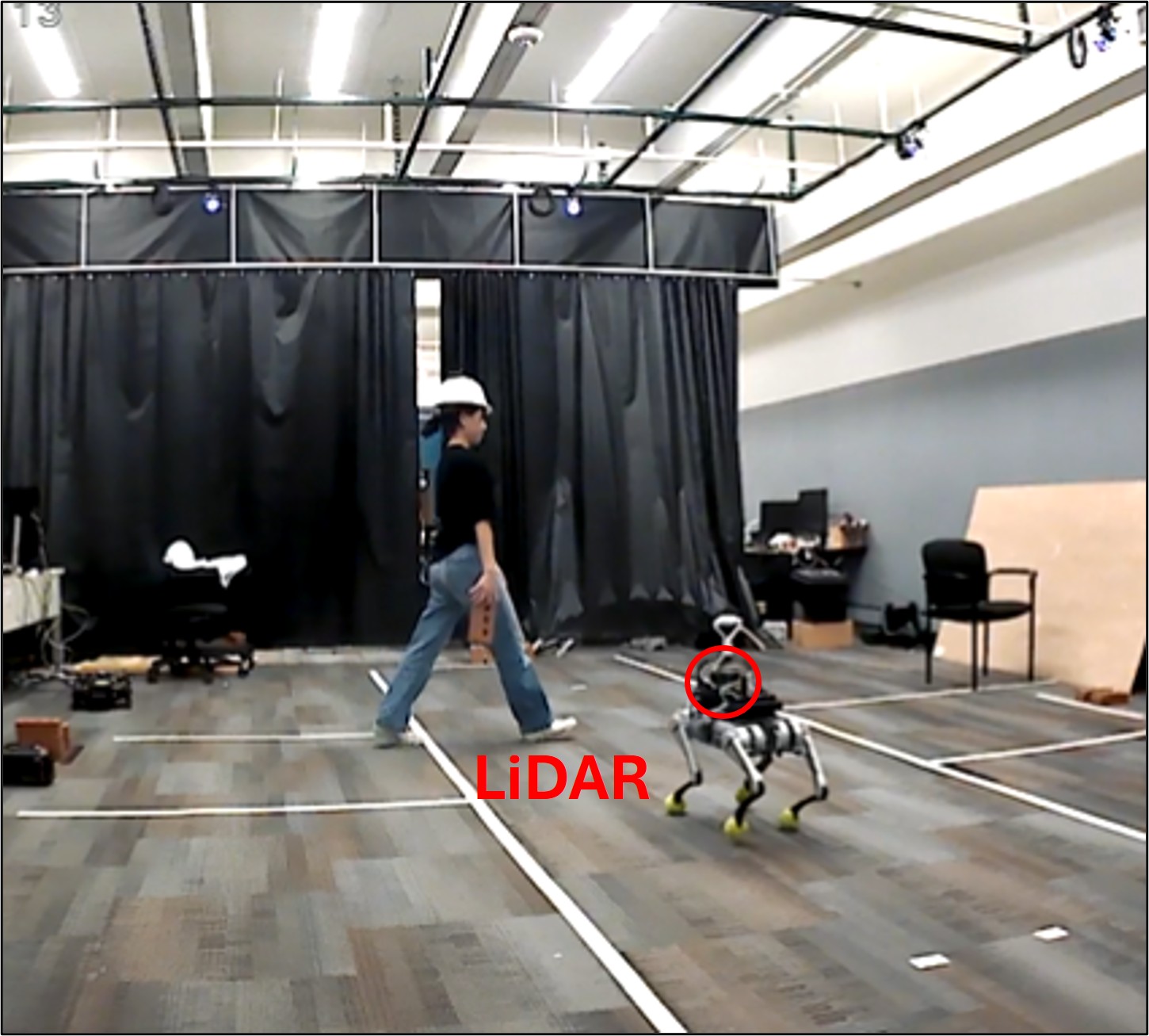}%
\label{fig_bricklaying_robot}}
\caption{Bricklaying testbed. (a) Layout. (b) Robot with LiDAR.}
\label{fig_bricklaying}
\end{figure}

{\bf{Dataset:}}
The experiment was conducted in accordance with the university's Institutional Review Board (H23040) and performed in a laboratory testbed \cite{kim2025understanding} configured to represent a low-density indoor construction site, as shown in Fig.~\ref{fig_bricklaying}. Within this workspace, the robot and a human participant carried out separate tasks concurrently: the robot collected environmental data while the participant constructed brick walls. The robotic platform consisted of a Unitree Go1 Edu quadruped equipped with a Velodyne LiDAR and onboard computing units. To generate representative human-robot interactions (HRIs), the robot was directed through a set of predefined waypoints designed to produce merging, passing, and crossing encounters with the participant. These scenarios emulate spatial conflicts commonly encountered in shared construction environments, where maintaining the original trajectories and velocities could result in unsafe or uncomfortably close interactions.

{\bf{Experiment:}} 
In previous work \cite{chen2027enhancing}, the repulsive forces associated with the human co-workers during a bricklaying task were modeled using the social force formulation \cite{Helbing_1995} to adjust local planner-generated waypoints. At each time step \(t\), the resulting force \(F_i(t)\) was applied as a displacement to the original path waypoint \(p_i(t)\), producing the adjusted waypoint \(p_i^{o}(t)\), as defined in \eqref{eqn_sfm_offset}. In this paper, we adopt the same social force model (SFM) to demonstrate the use of the proposed T-ConSG in dynamics-aware navigation instead of fused LiDAR-camera tracking \cite{chen2027enhancing}, with particular emphasis on incorporating dynamic-object nodes into trajectory adjustment. For the metrics, we use path length, path irregularity, closest pedestrian distance (CPD), and minimum time-to-collision (TTC) to characterize navigation efficiency, path smoothness, spatial clearance, and collision timing across 10 trials, respectively.
\begin{align}
p^{o}_{i}(t) &= p_i(t) + F_i(t) \label{eqn_sfm_offset}
\end{align}
{\bf{Discussion:}} 
As reported in Table~\ref{tab_navigation}, the SFM-adjusted trajectories achieve a similar mean closest pedestrian distance (\(2.62\,\mathrm{m}\) versus \(2.60\,\mathrm{m}\)), while the mean minimum time-to-collision rises from \(4.76\,\mathrm{s}\) to \(5.11\,\mathrm{s}\). These results indicate that incorporating the scene-graph-derived object states into the social force model generally produces trajectories with additional time for collision avoidance. The small change in closest pedestrian distance suggests that the application of SFM offsets \cite{chen2027enhancing} in the bricklaying dataset mainly influenced the timing and shape of the avoidance response rather than its ultimate spatial clearance. The adjustment trades path efficiency and smoothness for modest timing changes, and the experiment demonstrates the potential of integrating T-ConSG into an existing planner.
\begin{table}[!t]
\vspace*{6pt}
\caption{Evaluation results on the bricklaying test. Values are means ± standard deviations across 10 trials.}
\label{tab_navigation}
\centering
\begin{tabular}{|c|c|c|c|c|}
\hline
Planner & Path len. & Path irr. & CPD& Min. TTC\\
 & (m) & (rad/m)& (m)& (s)\\
\hline
Original & 83.13$\pm$8.12& 8.49$\pm$1.07& 2.60$\pm$0.10& 4.76$\pm$1.77\\
SFM & 88.48$\pm$8.81& 10.64$\pm$1.26& 2.62$\pm$0.10& 5.11$\pm$1.93\\
\hline
\end{tabular}
\end{table}
\subsection{Construction Progress Monitoring}
{\bf{Dataset:}}
The ConSLAM dataset \cite{trzeciak2023conslam} is a publicly available multimodal benchmark developed to evaluate SLAM algorithms and support progress monitoring in evolving construction environments. It comprises five sequences collected approximately once per month on the same floor of an active construction site, thereby capturing substantial geometric and structural changes as construction progressed. Data were acquired using a prototype handheld scanning system and include spatially registered and temporally synchronized LiDAR measurements and inertial observations. Its repeated surveys of the same evolving environment make ConSLAM particularly valuable for studying long-term change. In practice, we use the last four sequences for progress monitoring since the dataset maintainers flag Sequence 1 as faulty \cite{trzeciak2023conslam}. 

\begin{figure}[!htbp]
\vspace*{4pt}
\centering
\subfloat[]{\includegraphics[width=2.60in, trim={0 0.5mm 0 1mm}, clip]{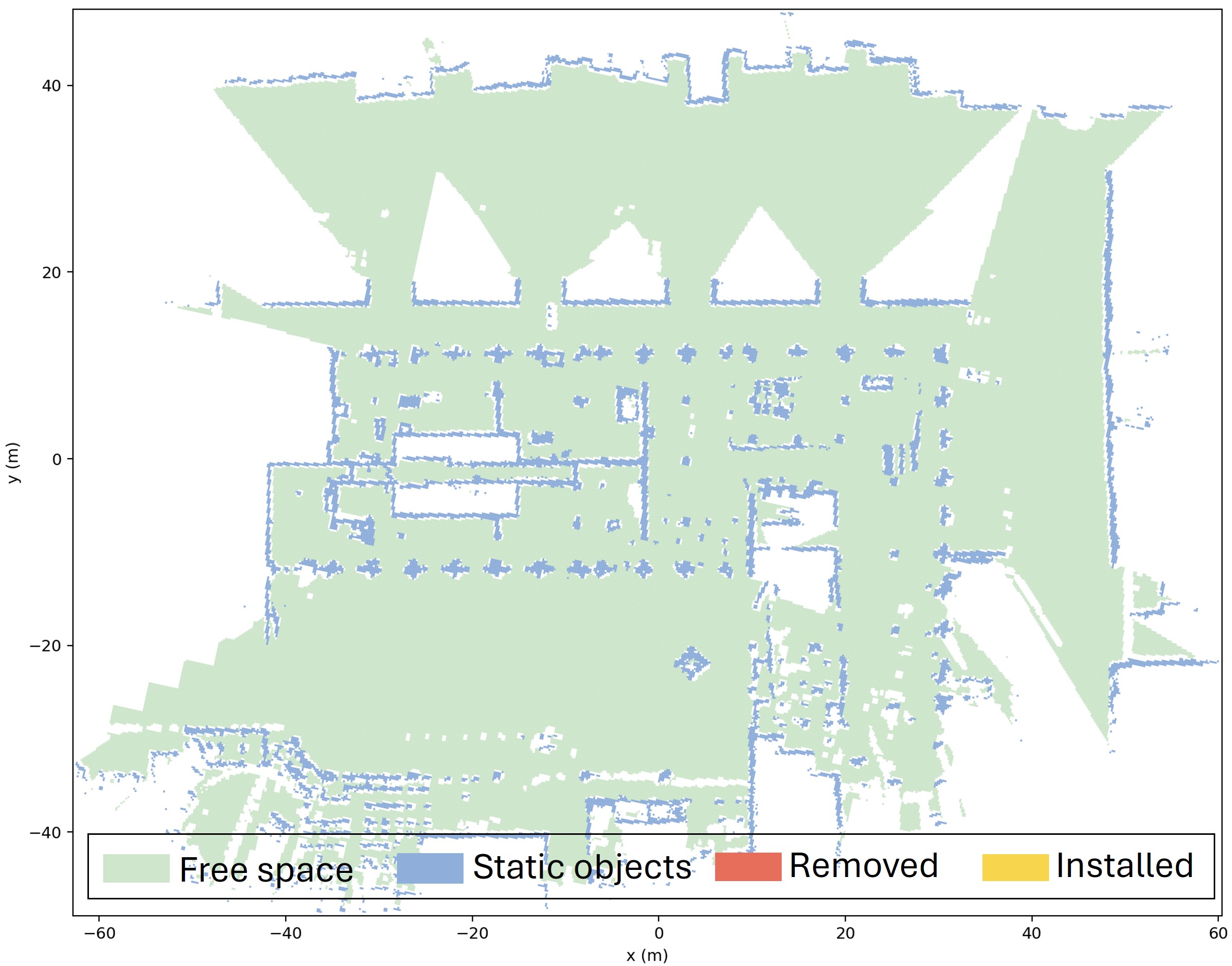}%
\label{fig_conslam_5}}\\[-2pt]
\subfloat[]{\includegraphics[width=2.60in, trim={0 1mm 0 1mm}, clip]{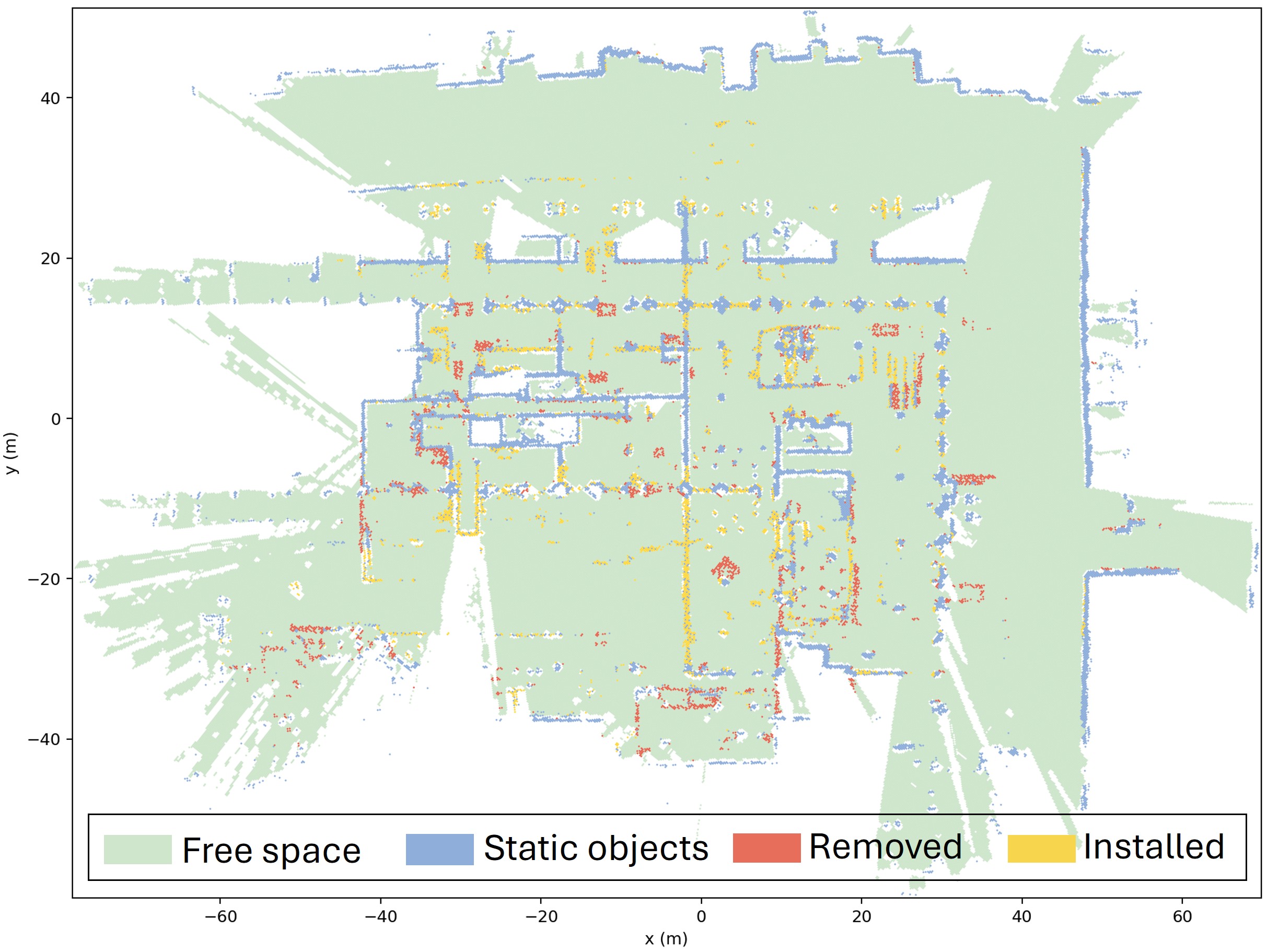}%
\label{fig_conslam_6}}\\[-2pt]
\subfloat[]{\includegraphics[width=2.60in, trim={0 1mm 0 1mm}, clip]{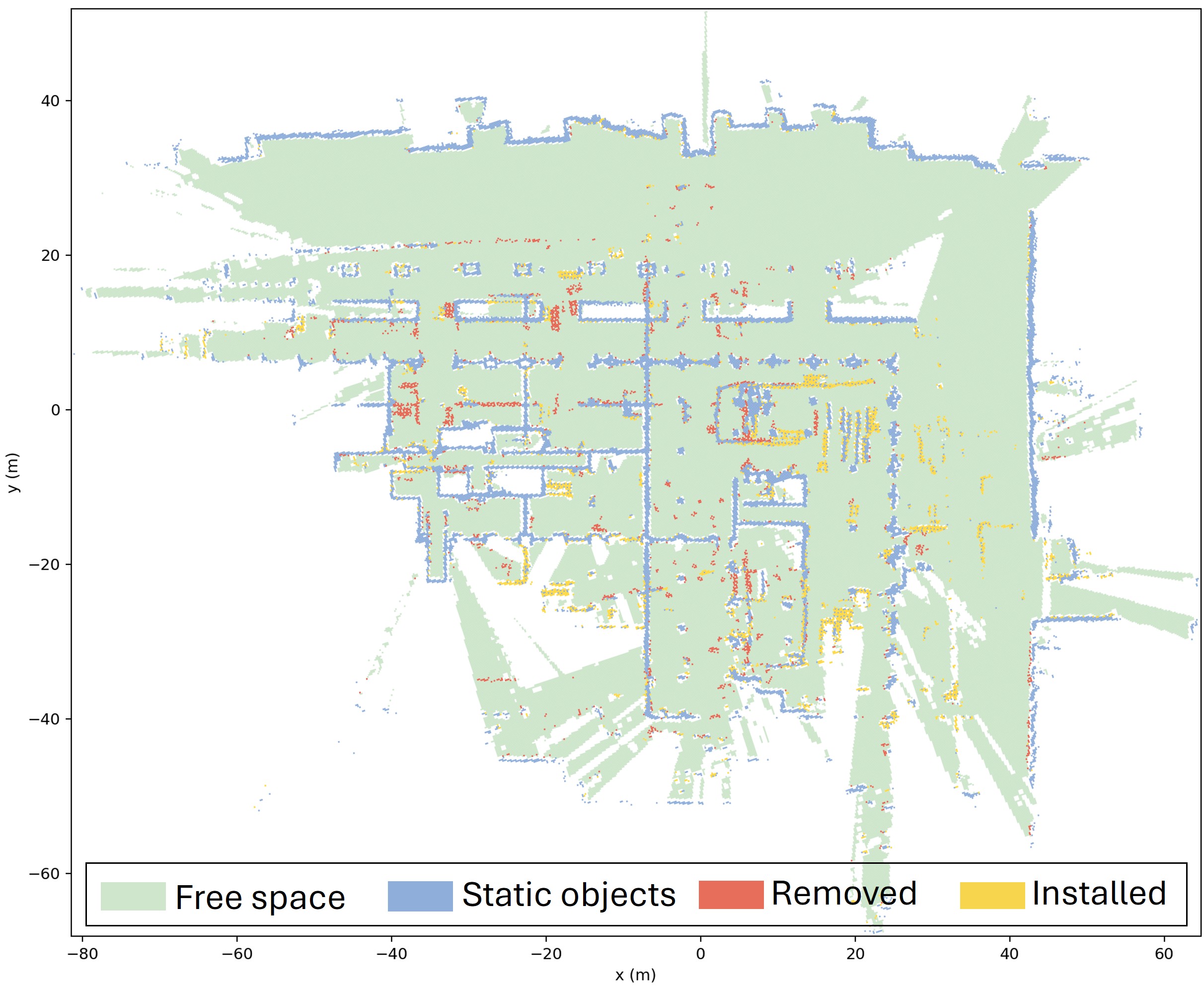}%
\label{fig_conslam_7}}\\[-2pt]
\subfloat[]{\includegraphics[width=2.60in, trim={0 1mm 0 1mm}, clip]{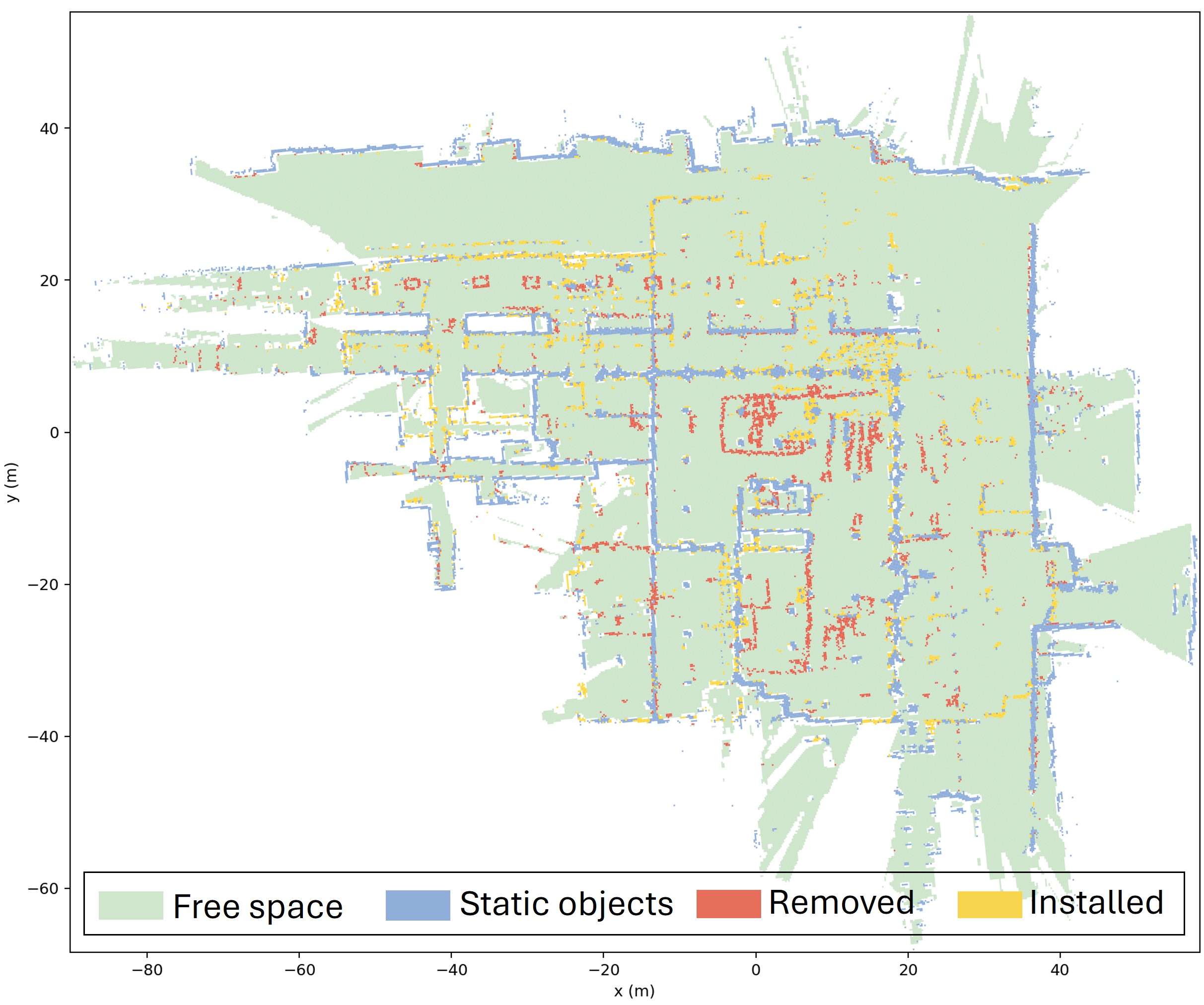}%
\label{fig_conslam_8}}\\[-2pt]
\caption{Free-space and static-object regions, together with estimated removal and installation regions across ConSLAM sequences: (a) Sequence 2. (b) Sequence 3. (c) Sequence 4. (d) Sequence 5.}
\label{fig_conslam_4dsg}
\end{figure}

{\bf{Experiment:}} 
To evaluate the proposed T-ConSG for construction progress monitoring, ConSLAM Sequences 2-5 were processed independently in chronological order. For each sequence, the free-space and static-object maps produced at the completion of the SLAM trajectory are retained as representations of the corresponding construction stage, as shown in Fig.~\ref{fig_conslam_4dsg}. The total areas occupied by free-space nodes and static-object nodes are then calculated to characterize the spatial evolution of the site across the four recording periods. Furthermore, we evaluated reference-region recovery using recall within the defined evaluation domain, as not all the observed structures along the SLAM trajectory are covered by the ground-truth point cloud. For Sequences 3-5, each map is compared with that of the preceding sequence to quantify two directional changes: \emph{installed} regions, which were previously observed as free space but are classified as static objects in the current sequence, and \emph{removed} regions, which were previously occupied by static objects but are subsequently observed as free. The resulting free-space area, static-object area, and inter-sequence transition areas are summarized in Table~\ref{tab_progress} to demonstrate how the graph representation can support quantitative monitoring of construction-related spatial changes over time.

\textbf{Discussion:}
As shown in Fig.~\ref{fig_conslam_4dsg} and Table~\ref{tab_progress}, the proposed representation consistently recovers most of the ground-truth free space across the four ConSLAM sequences, achieving recall values between \(89.82\%\) and \(92.81\%\). Static-object recall is more variable, potentially because of differences in observation coverage, ranging from \(76.77\%\) to \(93.51\%\). Despite this variation, the total mapped static-object area increases from \(540.24\,\mathrm{m}^2\) in Sequence~2 to \(854.64\,\mathrm{m}^2\) in Sequence~5, consistent with the progressive accumulation of structural elements as construction advances. The inter-sequence comparisons further identify changes in both directions: regions transitioning from free space to static occupancy are interpreted as installed areas, whereas static-to-free transitions indicate removed areas. These results illustrate that the free-space and static-object nodes can provide quantitative indicators of spatial evolution across construction stages.
\begin{table}[!t]
\vspace*{6pt}
\caption{Free-space and static-object areas and recall, with estimated inter-sequence installation and removal areas across ConSLAM sequences.}
\label{tab_progress}
\centering
\begin{tabular}{|c|c|c|c|c|c|c|}
\hline
Seq. & Free & Recall & Static & Recall& Installed & Removed\\
 & (\(m^2\)) & (\%) & (\(m^2\)) & (\%) & (\(m^2\)) & (\(m^2\))\\
\hline
2 & 5646.20 & 92.39 & 540.24 & 92.89 & -& -\\
3 & 7439.64 & 92.81 & 704.68 & 82.66 & 157.40& 100.28\\
4 & 5786.56 & 89.82 & 740.24 & 93.51 & 140.52& 105.70\\
5 & 6249.84 & 90.86 & 854.64 & 76.77 & 233.48& 173.60\\
\hline
\end{tabular}
\end{table}
\subsection{Online Change Detection}
\begin{figure}[!t]
\centering
\subfloat[]{\includegraphics[width=0.43\linewidth]{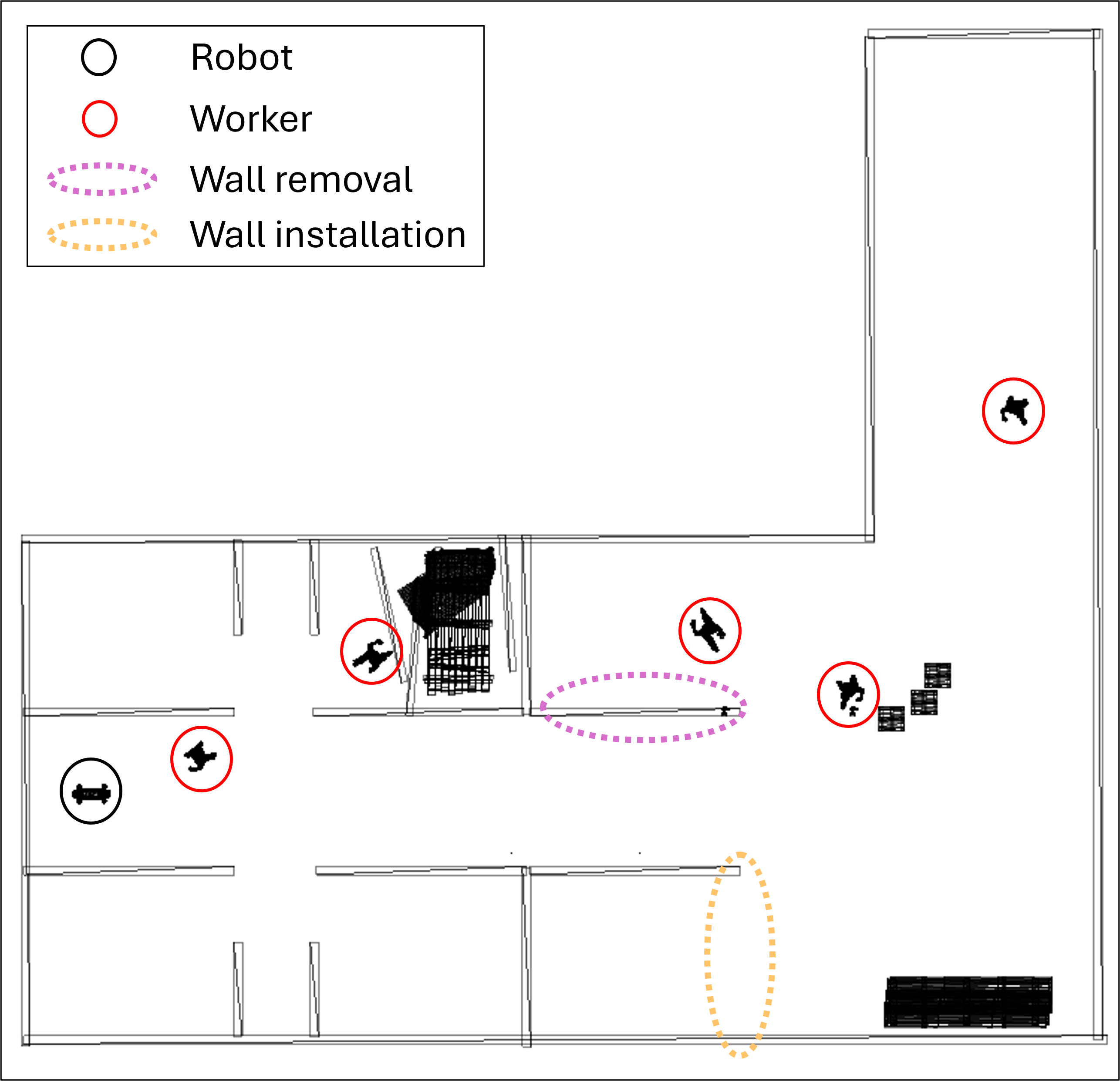}%
\label{fig_sim_layout}}
\subfloat[]{\includegraphics[width=0.53\linewidth]{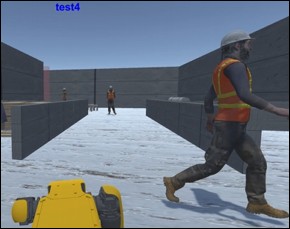}%
\label{fig_sim_unity}}
\caption{Simulated construction site. (a) Layout. (b) Unity view.}
\label{fig_sim}
\end{figure}

{\bf{Dataset:}}
The bricklaying dataset captures short-term dynamics, such as pedestrian motion, within individual recording sessions, whereas ConSLAM provides repeated observations of an evolving construction site and therefore represents longer-term environmental changes. However, neither dataset contains both short-term object motion and structural changes within a single continuous trial. To evaluate the event nodes of the proposed T-ConSG, we therefore employed a simulated construction environment developed in Unity (Fig.~\ref{fig_sim_unity}). This environment was originally created in previous work \cite{KIM2024105455} to investigate HRI patterns. For this study, a simulated Velodyne LiDAR and an inertial measurement unit (IMU) were mounted on a quadruped robot to generate ROS-compatible sensor data for the SLAM pipeline. In addition to the existing moving construction workers, the environment was extended with 5 trials each contained one installation and one removal event  (Fig.~\ref{fig_sim_layout}). Both structural changes were configured to occur outside the robot's field of view, requiring the system to identify them when the affected areas were subsequently revisited.

\begin{figure}[!t]
\centering
\subfloat[]{\includegraphics[width=0.48\linewidth]{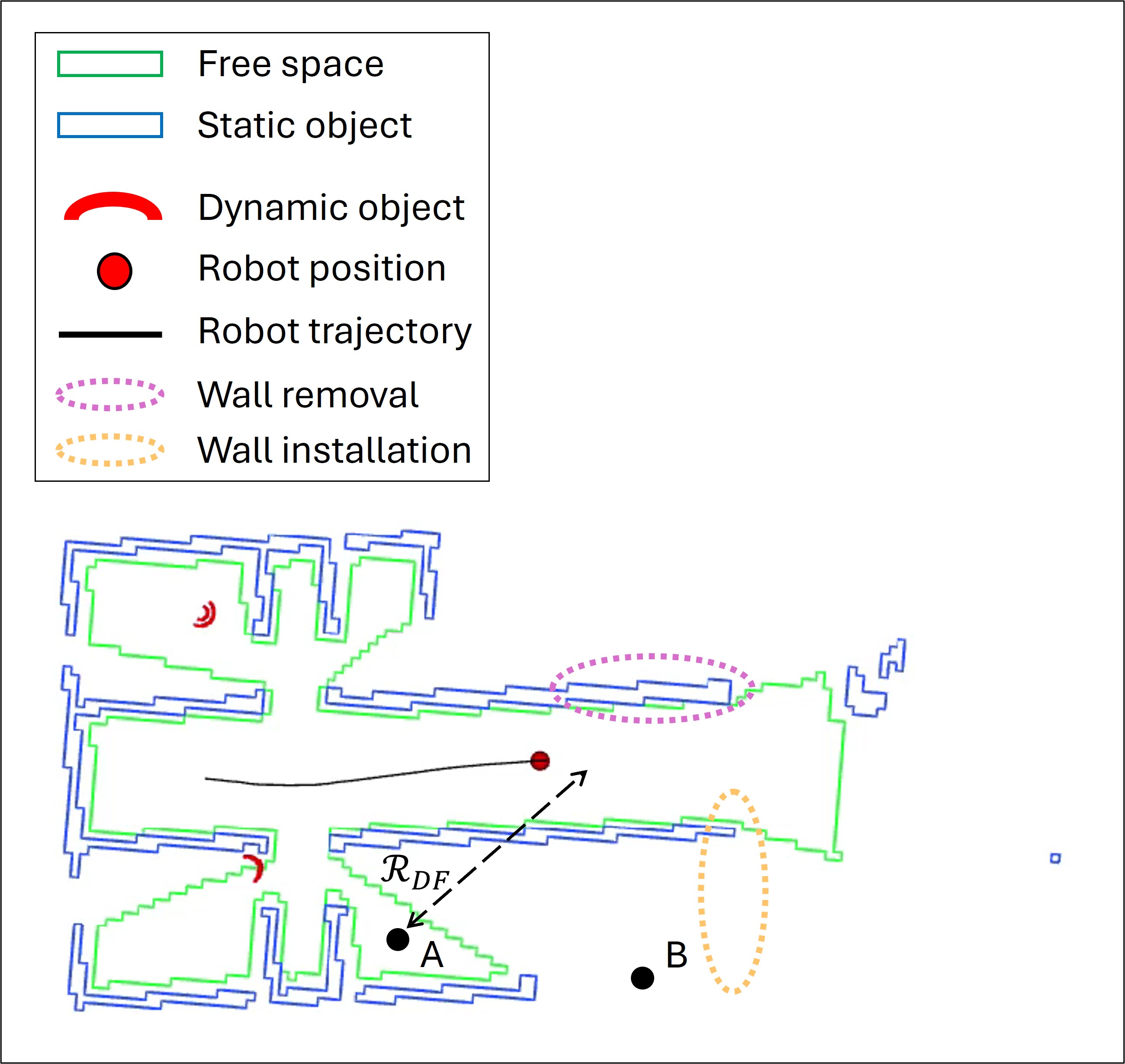}%
\label{fig_sim_4dsg_1}}
\subfloat[]{\includegraphics[width=0.48\linewidth]{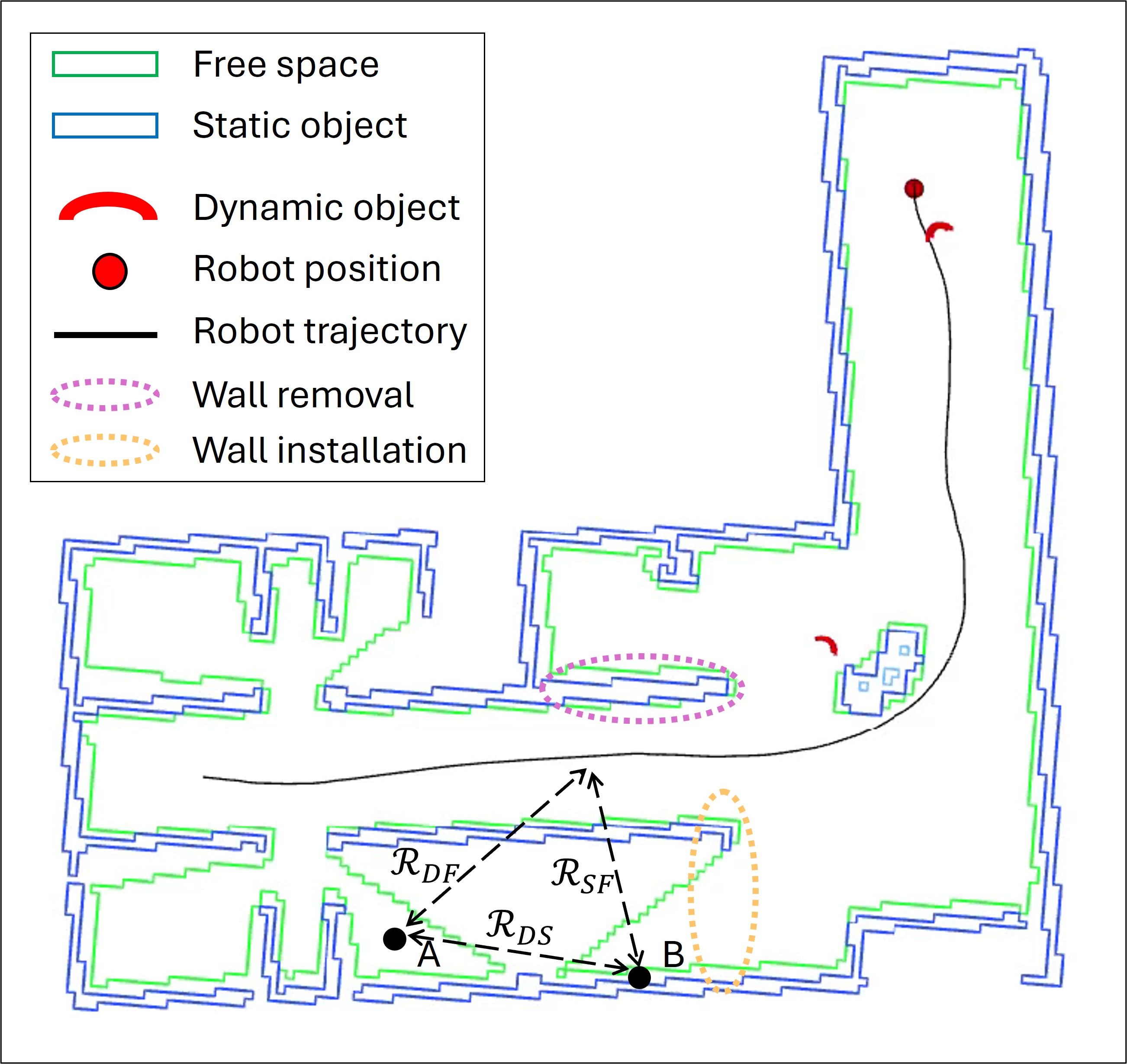}%
\label{fig_sim_4dsg_2}}
\hfil
\subfloat[]{\includegraphics[width=0.48\linewidth]{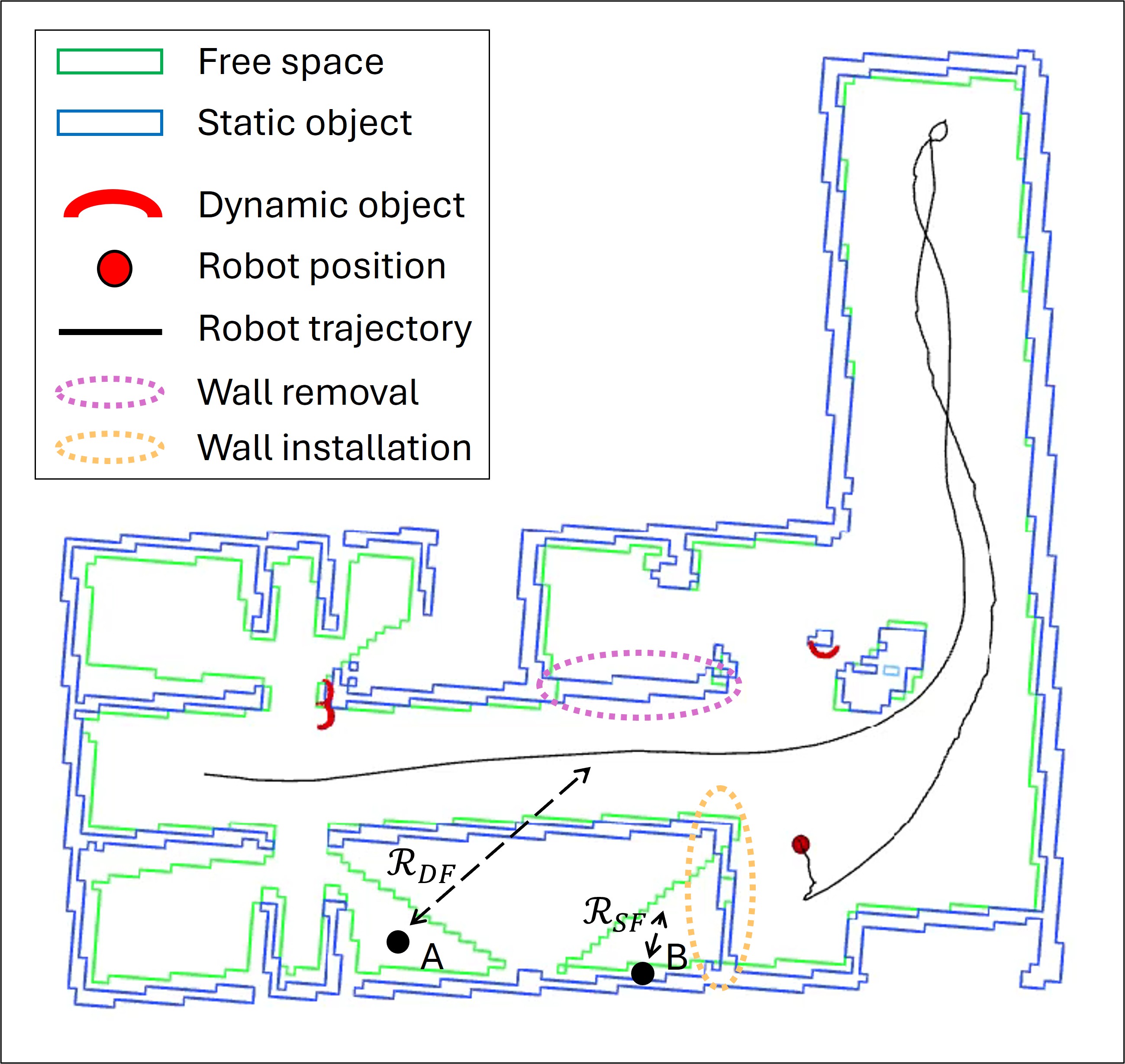}%
\label{fig_sim_4dsg_3}}
\subfloat[]{\includegraphics[width=0.48\linewidth]{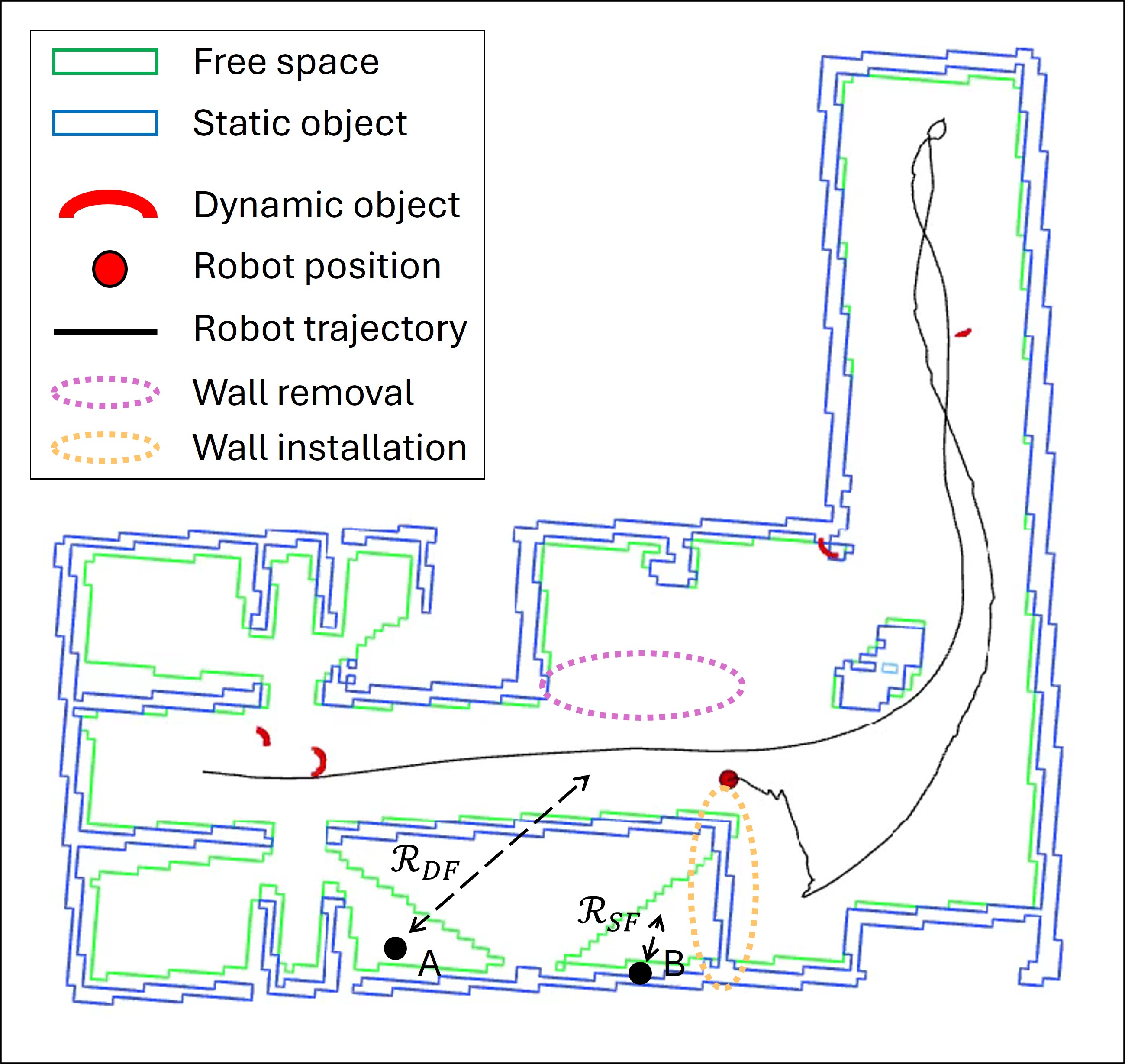}%
\label{fig_sim_4dsg_4}}
\caption{Free space and static object maps along the robot trajectory in trial 1. The free space, static objects, dynamic objects, and spatial-relational edges related to A and B are visualized before and after the structural changes, including the time when the changes occur. (a) Initial exploration. (b) Structural changes occur. (c) Installation detected. (d) Removal detected.}
\label{fig_sim_4dsg}
\end{figure}

{\bf{Experiment:}} 
Fig.~\ref{fig_sim_4dsg} illustrates the trajectory followed by the quadruped robot during one trial. Starting from its initial position, the robot first traversed a narrow corridor while encountering nearby moving workers (Fig.~\ref{fig_sim_4dsg_1}). It then navigated around a group of boxes and proceeded toward the upper-right region of the map. Upon reaching a predefined location (Fig.~\ref{fig_sim_4dsg_2}), the simulation triggered two structural changes outside the robot's current field of view: a wall segment on the left side of the corridor was removed, while another wall segment was installed on the right side. The robot subsequently returned to inspect the newly installed wall (Fig.~\ref{fig_sim_4dsg_3}), passed through the space previously occupied by the removed wall (Fig.~\ref{fig_sim_4dsg_4}), and finally arrived back at its starting position. Throughout the trial, the free-space and static-object nodes continuously maintained and published their persistent maps, which were updated by confirmed events. The event node examined spatial conflicts between these historical representations and classified each conflicting region as an installation or removal event according to the current TSDF evidence. To demonstrate how structural changes affect graph-based accessibility, we query whether the robot from point \(A\), located within a free-space node, can access point \(B\) through the spatial-relation edge \(\mathcal{R}_{SF}=\texttt{connected}\) between that free-space node and the static-object node containing \(B\).

{\bf{Discussion:}}
The results in Fig.~\ref{fig_sim_4dsg} demonstrate that the proposed T-ConSG can identify structural changes that occur outside the robot's field of view. At the beginning of the trial, the free-space and static-object nodes incrementally construct their respective maps as the robot explores the environment. Although the wall installation and removal occur while the affected regions are unobserved, the changes are detected when the robot later revisits these locations. As shown in Fig.~\ref{fig_sim_4dsg_3}, the newly installed wall produces a conflict with the historically observed free-space representation, and the current TSDF evidence enables the event node to classify this conflict as an installation. Conversely, when the robot traverses the area previously occupied by the removed wall, the observed free-space evidence conflicts with the accumulated static-object map, resulting in the removal event shown in Fig.~\ref{fig_sim_4dsg_4}. Meanwhile, nearby moving workers are represented as dynamic objects and do not produce persistent structural-change events, indicating that the graph preserves useful historical knowledge while distinguishing short-term object motion from persistent modifications to the environment. In total, all 10 structural changes in 5 trials were correctly classified without false event detections. The accessibility query also changes as the graph is updated. In Fig.~\ref{fig_sim_4dsg_1}, point \(B\) is not associated with any static-object node; therefore, no \(\mathcal{R}_{SF}\) edge supports a graph-inferred access relation from \(A\) to \(B\), whereas in Fig.~\ref{fig_sim_4dsg_2}, the static-object node containing \(B\) is connected to the free-space node containing \(A\), establishing \(\mathcal{R}_{SF}=\texttt{connected}\), \(\mathcal{R}_{DF} = \texttt{inside}\) and \(\mathcal{R}_{DS}=\texttt{can\_access}\), indicating that the robot can access \(B\). In Fig.~\ref{fig_sim_4dsg_3}, the detected structural change disconnects this static-object node from the free-space node, removing the corresponding \(\mathcal{R}_{SF}\) edge and the inferred accessibility.

\section{Conclusion and Future Work}
This work presented a LiDAR-based real-time temporal construction scene graph for representing dynamic and evolving environments. The proposed framework integrates persistent spatial evidence with explicit environmental events, supporting reasoning over both transient dynamics and persistent changes. Experiments also demonstrate its applicability to evolving construction environments. However, several limitations still remain. First, the current implementation uses 2D spatial representations; future work will extend the framework to real-world 3D environments. Second, apart from the current experiments, future work will more explicitly exploit free-space regions for accessibility reasoning and temporal change querying. Finally, incorporating semantic information could additionally enable detected geometric changes to be associated with construction elements and activities.
\bibliographystyle{IEEEtran}
\bibliography{cas-refs}

@article{KIM2024105455,
title = {Understanding human-robot proxemic norms in construction: How do humans navigate around robots?},
journal = {Automation in Construction},
volume = {164},
pages = {105455},
year = {2024},
issn = {0926-5805},
doi = {10.1016/j.autcon.2024.105455},
url = {https://www.sciencedirect.com/science/article/pii/S0926580524001912},
author = {YeSeul Kim and Seongyong Kim and Yilong Chen and HyunJin Yang and Seungwoo Kim and Sehoon Ha and Matthew Gombolay and Yonghan Ahn and Yong Kwon Cho},
}

@article{kim2025understanding,
  title={Understanding the effects of humanlike robot motions on unfocused human--robot interaction},
  author={Kim, YeSeul and Chen, Yilong and Gombolay, Mathew and Cho, Yong K},
  journal={Advanced Engineering Informatics},
  volume={68},
  pages={103751},
  year={2025},
  publisher={Elsevier}
}

@article{chen2027enhancing,
  title={Enhancing Socially Aware Construction Robot Navigation through LiDAR-Camera Sensor Fusion and Dynamic Object Tracking},
  author={Chen, Yilong and Kwon Cho, Yong},
  journal={Journal of Computing in Civil Engineering},
  volume={41},
  number={1},
  pages={04026113},
  year={2027},
  publisher={American Society of Civil Engineers}
}

@article{Schmid_2023,
   title={Dynablox: Real-Time Detection of Diverse Dynamic Objects in Complex Environments},
   volume={8},
   ISSN={2377-3774},
   url={http://dx.doi.org/10.1109/LRA.2023.3305239},
   DOI={10.1109/lra.2023.3305239},
   number={10},
   journal={IEEE Robotics and Automation Letters},
   publisher={Institute of Electrical and Electronics Engineers (IEEE)},
   author={Schmid, Lukas and Andersson, Olov and Sulser, Aurelio and Pfreundschuh, Patrick and Siegwart, Roland},
   year={2023},
   month=Oct, pages={6259-6266} }

@article{Helbing_1995,
   title={Social force model for pedestrian dynamics},
   volume={51},
   ISSN={1095-3787},
   url={http://dx.doi.org/10.1103/PhysRevE.51.4282},
   DOI={10.1103/physreve.51.4282},
   number={5},
   journal={Physical Review E},
   publisher={American Physical Society (APS)},
   author={Helbing, Dirk and Molnár, Péter},
   year={1995},
   month=May, pages={4282-4286} }

@inproceedings{armeni20193d,
  title={3d scene graph: A structure for unified semantics, 3d space, and camera},
  author={Armeni, Iro and He, Zhi-Yang and Gwak, JunYoung and Zamir, Amir R and Fischer, Martin and Malik, Jitendra and Savarese, Silvio},
  booktitle={Proceedings of the IEEE/CVF international conference on computer vision},
  pages={5664--5673},
  year={2019}
}

@article{rosinol20203d,
  title={3D dynamic scene graphs: Actionable spatial perception with places, objects, and humans},
  author={Rosinol, Antoni and Gupta, Arjun and Abate, Marcus and Shi, Jingnan and Carlone, Luca},
  journal={arXiv preprint arXiv:2002.06289},
  year={2020}
}

@article{hughes2022hydra,
  title={Hydra: A real-time spatial perception system for 3D scene graph construction and optimization},
  author={Hughes, Nathan and Chang, Yun and Carlone, Luca},
  journal={arXiv preprint arXiv:2201.13360},
  year={2022}
}

@inproceedings{chang2023hydra,
  title={Hydra-multi: Collaborative online construction of 3d scene graphs with multi-robot teams},
  author={Chang, Yun and Hughes, Nathan and Ray, Aaron and Carlone, Luca},
  booktitle={2023 IEEE/RSJ International Conference on Intelligent Robots and Systems (IROS)},
  pages={10995--11002},
  year={2023},
  organization={IEEE}
}

@inproceedings{greve2024collaborative,
  title={Collaborative dynamic 3d scene graphs for automated driving},
  author={Greve, Elias and B{\"u}chner, Martin and V{\"o}disch, Niclas and Burgard, Wolfram and Valada, Abhinav},
  booktitle={2024 IEEE International Conference on Robotics and Automation (ICRA)},
  pages={11118--11124},
  year={2024},
  organization={IEEE}
}

@article{schmid2024khronos,
  title={Khronos: A unified approach for spatio-temporal metric-semantic slam in dynamic environments},
  author={Schmid, Lukas and Abate, Marcus and Chang, Yun and Carlone, Luca},
  journal={arXiv preprint arXiv:2402.13817},
  year={2024}
}

@article{rotondi20263d,
  title={3D Scene Graphs: Open Challenges and Future Directions},
  author={Rotondi, Dennis and Argenziano, Francesco and Koch, Sebastian and Hughes, Nathan and Buechner, Martin and Wald, Johanna and Schmid, Lukas Rosenberger and Nardi, Daniele and Valada, Abhinav and Paull, Liam and others},
  journal={arXiv preprint arXiv:2606.19383},
  year={2026}
}

@article{wang2023trajectory,
  title={Trajectory distribution aware graph convolutional network for trajectory prediction considering spatio-temporal interactions and scene information},
  author={Wang, Ruiping and Hu, Zhijian and Song, Xiao and Li, Wenxin},
  journal={IEEE Transactions on Knowledge and Data Engineering},
  volume={36},
  number={8},
  pages={4304--4316},
  year={2023},
  publisher={IEEE}
}

@inproceedings{gorlo2026describe,
  title={Describe anything anywhere at any moment},
  author={Gorlo, Nicolas and Schmid, Lukas and Carlone, Luca},
  booktitle={Proceedings of the IEEE/CVF Conference on Computer Vision and Pattern Recognition},
  pages={35002--35013},
  year={2026}
}

@inproceedings{nguyen2025react,
  title={REACT: Real-time efficient attribute clustering and transfer for updatable 3D scene graph},
  author={Nguyen, Phuoc and Verdoja, Francesco and Kyrki, Ville},
  booktitle={2025 IEEE/RSJ International Conference on Intelligent Robots and Systems (IROS)},
  pages={2209--2216},
  year={2025},
  organization={IEEE}
}

@article{rosinol2021kimera,
  title={Kimera: From SLAM to spatial perception with 3D dynamic scene graphs},
  author={Rosinol, Antoni and Violette, Andrew and Abate, Marcus and Hughes, Nathan and Chang, Yun and Shi, Jingnan and Gupta, Arjun and Carlone, Luca},
  journal={The International Journal of Robotics Research},
  volume={40},
  number={12-14},
  pages={1510--1546},
  year={2021},
  publisher={SAGE Publications Sage UK: London, England}
}

@incollection{sh-ch15-dyndef,
  title        = {Dynamic and Deformable {SLAM}},
  author       = {Lukas Schmid and Jos{\'e} Mar{\'i}a Mart{\'i}nez Montiel and Shoudong Huang and Daniel Cremers and Jos{\'e} Neira and Javier Civera},
  booktitle    = {{SLAM Handbook.} From Localization and Mapping to Spatial Intelligence},
  publisher    = {Cambridge University Press},
  editor       = {Luca Carlone and Ayoung Kim and Timothy Barfoot and Daniel Cremers and Frank Dellaert},
  year         = {2026}
}

@inproceedings{looper20233d,
  title={3d vsg: Long-term semantic scene change prediction through 3d variable scene graphs},
  author={Looper, Samuel and Rodriguez-Puigvert, Javier and Siegwart, Roland and Cadena, Cesar and Schmid, Lukas},
  booktitle={2023 IEEE International Conference on Robotics and Automation (ICRA)},
  pages={8179--8186},
  year={2023},
  organization={IEEE}
}

@inproceedings{judd2018multimotion,
  title={Multimotion visual odometry (MVO): Simultaneous estimation of camera and third-party motions},
  author={Judd, Kevin M and Gammell, Jonathan D and Newman, Paul},
  booktitle={2018 IEEE/RSJ International Conference on Intelligent Robots and Systems (IROS)},
  pages={3949--3956},
  year={2018},
  organization={IEEE}
}

@article{zhang2020vdo,
  title={VDO-SLAM: A visual dynamic object-aware SLAM system},
  author={Zhang, Jun and Henein, Mina and Mahony, Robert and Ila, Viorela},
  journal={arXiv preprint arXiv:2005.11052},
  year={2020}
}

@article{bescos2021dynaslam,
  title={DynaSLAM II: Tightly-coupled multi-object tracking and SLAM},
  author={Bescos, Berta and Campos, Carlos and Tard{\'o}s, Juan D and Neira, Jos{\'e}},
  journal={IEEE robotics and automation letters},
  volume={6},
  number={3},
  pages={5191--5198},
  year={2021},
  publisher={IEEE}
}

@inproceedings{runz2017co,
  title={Co-fusion: Real-time segmentation, tracking and fusion of multiple objects},
  author={R{\"u}nz, Martin and Agapito, Lourdes},
  booktitle={2017 IEEE International Conference on Robotics and Automation (ICRA)},
  pages={4471--4478},
  year={2017},
  organization={IEEE}
}

@article{sun2025nothing,
  title={Nothing stands still: A spatiotemporal benchmark on 3d point cloud registration under large geometric and temporal change},
  author={Sun, Tao and Hao, Yan and Huang, Shengyu and Savarese, Silvio and Schindler, Konrad and Pollefeys, Marc and Armeni, Iro},
  journal={ISPRS Journal of Photogrammetry and Remote Sensing},
  volume={220},
  pages={799--823},
  year={2025},
  publisher={Elsevier}
}

@inproceedings{rowell2024lista,
  title={LiSTA: Geometric object-based change detection in cluttered environments},
  author={Rowell, Joseph and Zhang, Lintong and Fallon, Maurice},
  booktitle={2024 IEEE International Conference on Robotics and Automation (ICRA)},
  pages={3632--3638},
  year={2024},
  organization={IEEE}
}

@inproceedings{fehr2017tsdf,
  title={TSDF-based change detection for consistent long-term dense reconstruction and dynamic object discovery},
  author={Fehr, Marius and Furrer, Fadri and Dryanovski, Ivan and Sturm, J{\"u}rgen and Gilitschenski, Igor and Siegwart, Roland and Cadena, Cesar},
  booktitle={2017 IEEE International Conference on Robotics and automation (ICRA)},
  pages={5237--5244},
  year={2017},
  organization={IEEE}
}

@inproceedings{schmid2022panoptic,
  title={Panoptic multi-tsdfs: a flexible representation for online multi-resolution volumetric mapping and long-term dynamic scene consistency},
  author={Schmid, Lukas and Delmerico, Jeffrey and Sch{\"o}nberger, Johannes L and Nieto, Juan and Pollefeys, Marc and Siegwart, Roland and Cadena, Cesar},
  booktitle={2022 International Conference on Robotics and Automation (ICRA)},
  pages={8018--8024},
  year={2022},
  organization={IEEE}
}

@article{trzeciak2023conslam,
  title={ConSLAM: Construction data set for SLAM},
  author={Trzeciak, Maciej and Pluta, Kacper and Fathy, Yasmin and Alcalde, Lucio and Chee, Stanley and Bromley, Antony and Brilakis, Ioannis and Alliez, Pierre},
  journal={Journal of Computing in Civil Engineering},
  volume={37},
  number={3},
  pages={04023009},
  year={2023},
  publisher={American Society of Civil Engineers}
}

@article{chen2022direct,
  title={Direct lidar-inertial odometry: Lightweight lio with continuous-time motion correction},
  author={Chen, Kenny and Nemiroff, Ryan and Lopez, Brett T},
  journal={arXiv preprint arXiv:2203.03749},
  year={2022}
}

@article{parascho2023construction,
  title={Construction robotics: From automation to collaboration},
  author={Parascho, Stefana},
  journal={Annual review of control, robotics, and autonomous systems},
  volume={6},
  number={1},
  pages={183--204},
  year={2023},
  publisher={Annual Reviews}
}

@article{chen2026localization,
  title={Localization and navigation of construction robots: A human-centric review},
  author={Chen, Junyang and Zhao, Zhiheng and Huang, George Q},
  journal={Automation in Construction},
  volume={187},
  pages={106953},
  year={2026},
  publisher={Elsevier}
}

@article{you2025construction,
  title={Construction robotics in extreme environments: From earth to space},
  author={You, Ke and Zhou, Cheng and Ding, Lieyun and Wang, Yuxiang},
  journal={Engineering},
  year={2025},
  publisher={Elsevier}
}
\newpage
\vfill
\end{document}